\documentclass[11pt,a4paper]{article}
\usepackage[hyperref]{acl2021}
\usepackage{times}
\usepackage{latexsym}

\usepackage{booktabs}
\usepackage{multirow}
\usepackage{amsmath}
\usepackage{graphicx}

\usepackage{microtype}

\aclfinalcopy

\title{EDIN: An End-to-end Benchmark and Pipeline for Unknown Entity Discovery and Indexing}

\author{Nora Kassner\textsuperscript{1,2},  Fabio Petroni\textsuperscript{1}, Mikhail Plekhanov\textsuperscript{1}, Sebastian Riedel\textsuperscript{1}, Nicola Cancedda\textsuperscript{1} \\
\textsuperscript{1}Meta AI \\
\textsuperscript{2}Center for Information and Language Processing, 
LMU Munich\\
\texttt{kassner@cis.lmu.de} 
}

\date{}

\begin{document}
\maketitle
\begin{abstract}
Existing work on Entity Linking mostly assumes that the reference knowledge base is complete, and therefore all mentions can be linked. In practice this is hardly ever the case, as knowledge bases are incomplete and because novel concepts arise constantly. This paper created the \emph{Unknown Entity Discovery and Indexing} (EDIN) benchmark where unknown entities, that is entities without a description in the knowledge base and labeled mentions,
have to be integrated into an existing entity linking system.  By contrasting EDIN with zero-shot entity linking, we provide insight on the additional challenges it poses.  Building on dense-retrieval based entity linking, we introduce the end-to-end EDIN pipeline that detects, clusters, and indexes mentions of unknown entities in context. Experiments show that indexing a single embedding per entity unifying the information of multiple mentions works better than indexing mentions independently.
\end{abstract}

\section{Introduction}
Entity linking (EL), the task of detecting mentions of entities in context and disambiguating them to a reference knowledge base (KB), is an  essential task for text understanding. Existing works on EL mostly assume that the reference KB is complete, and therefore all mentions can be linked. In practice this is hardly ever the case, as knowledge bases are incomplete and because novel concepts arise constantly. For example, English Wikipedia, which is often used as the reference KB for large scale linking, is growing by more than 17k entities every month.\footnote{\url{https://en.wikipedia.org/wiki/Wikipedia:Size_of_Wikipedia} (09.05.2022)}

We created the EDIN benchmark and pipeline\footnote{Model and benchmark are currently not publicly available.} where \emph{unknown entities}, that is entities with no available canonical names, descriptions and labeled mentions, have to be integrated into an existing EL model in an end-to-end fashion.

The benchmark is based on Wikipedia and a subset of news pages of the common-crawl dataset OSCAR \cite{AbadjiOrtizSuarezRomaryetal.2021}, split into two parts, one preceding time $t_1$ and one between $t_1$ and $t_2$. With current approaches, an EL system created at $t_{1}$ is unable to successfully link unknown entities that were added to Wikipedia between $t_{1}$ and a subsequent time $t_{2}$, as it lacks the ability to represent them as part of the entity index. This sets this task apart from zero-shot entity linking \cite{logeswaran-etal-2019-zero}. In the zero-shot (zs) setting, a textual description of the zs entities is assumed available at the time of training. This allows the creation of entity embeddings for them, their insertion in a dense entity index, and their use during training as negative examples. In contrast, no such textual description is initially available for unknown entities.

To adapt the model trained at $t_{1}$, the model can only make use of an \emph{adaptation dataset} and unsupervised techniques. There are two parts to this task: i) Discovery: The entity linking system needs to detect mentions of unknown entities part of the adaptation dataset and classify them as unknown and ii) Indexing: co-referring mentions of unknown entities need to be mapped to a single embedding compatible with the entity index. The EDIN-pipeline developed for this task is the first to adapt a dense-retrieval based EL system created at $t_{1}$ such that it incorporates unknown entities in an end-to-end fashion.

We show that distinguishing known from unknown entities, arguably a key feature of an intelligent system, poses a major challenge to dense-retrieval based EL systems. To successfully do so, a model has to strike the right balance between relying on mention vs. context. On one hand, the model needs to distinguish unknown entities carrying the same name as known entities and co-refer different mentions of the same unknown entities. In both cases, context reliance is necessary. E.g, the model needs to distinguish the 2019 novel `Underland' from multiple other novels and fictional references carrying the same name. On the other hand, the model needs to distinguish unknown entities with new names but semantic similarity. Here, mention reliance is important. E.g., the model needs to distinguishing BioNTech from other biotechnology companies with similar names and contexts.

Another challenge in discovery is class imbalance between known and unknown entities. With Wikipedia-size KBs, there tend to be many fewer mentions of unknown entities than of known entities and an adaptation dataset of finite size limits recall. In this work, we optimize for recall on the cost of precision.

On the side of indexing, inserting unknown entities into a space of known entities poses problems of interference with known entities in their close proximity. Again, consider the BioNTech example from above. Semantically, we want to place BioNTech in proximity of other biotech companies but in a way that dense-retrieval linking can still differentiate between them.
We find that re-training the model after indexing, and therefore giving it the chance to learn from hard negatives in the style of zs entity linking, is crucial to overcome this challenge. 

We experiment with different indexing methods. In particular, we contrast single mention-level indexing \citet{fitzgerald-etal-2021-moleman} with indexing clusters of mentions. We find that unifying the information of multiple mentions into a single embedding is beneficial.

By introducing a clear-cut temporal segmentation this benchmark targets unknown entities which are truly novel/unseen to all parts of an EL system, specifically including the pre-trained language model (PLM). Therefore, the EL system cannot rely on implicit knowledge captured by the PLM. This is, to the best of our knowledge, a setting that has not been explored before in the context of dense-retrieval based EL.

Temporal segmentation also lets us study the effect of entity encoder and PLM degradation. We note that EL precision drops for known entities in novel contexts which points to a large problem of PLM staleness also discussed by \cite{TACL3539, NEURIPS2021_f5bf0ba0}. Novel entities appearing in the context around known entities also complicate discovery, making it likely that the known entities will be mistakenly classified as unknown.

We summarize our contributions as follows:
i) We created the EDIN-benchmark, a large scale end-to-end entity linking benchmark dataset where unknown entities need to be discovered and integrated into an existing entity index. ii) We contrast this task with zero-shot entity linking, and provide insight on the additional challenges it poses. iii) We propose the EDIN-pipeline in the form of an extension of existing dense-retrieval architectures. vi) We experiment with different indexing methods, specifically indexing single mentions vs. clusters of mentions.

\section{Task definition}
We formally define end-to-end EL as follows: Given a paragraph $p$ and a set of known entities
$E_{K} = \{e_{i}\}$ from Wikipedia, each with canonical name, the title, $t(e_{i})$
and textual description $d(e_{i})$, our goal is to output a list of tuples, $(e, [i, j])$, where $e \in E_{K}$ is the entity corresponding to the mention $m_{i,j}$ spanning from the $i^{th}$ to $j^{th}$ token in $p$. We call a system that solves this task based on $d(e_{i})$ a \textit{Description-based} entity linking system L.

For EDIN-benchmark, after training a model $L_{t1}$ at time step $t_{1}$, a set of unknown entities $E_{U} = \{e_{i}\}$ with $ E_{U} \bigcap E_{K} = \emptyset$ and no available canonical name, description and labeled mentions is introduced between $t_{1}$ and $t_{2}$, with $t_{2}>t_{1}$. The task is to adapt $L_{t1}$ such that it can successfully link mentions of the union of $E_{U} \bigcup E_{K}$. 

We use three dataset splits: the training set $D_{train}$ to train $L_{t1}$, the adaptation dataset $D_{adapt}$ used to adapt $L_{t1}$ and the testset $D_{test}$ to evaluate. Both $D_{adapt}$ and $D_{test}$ include mentions between $t_{1}$ and $t_{2}$ but are disjoint, e.g., $D_{adapt} \bigcap D_{test} = \emptyset$.

\begin{figure*}[h!]
\centering
\includegraphics[width=\textwidth]{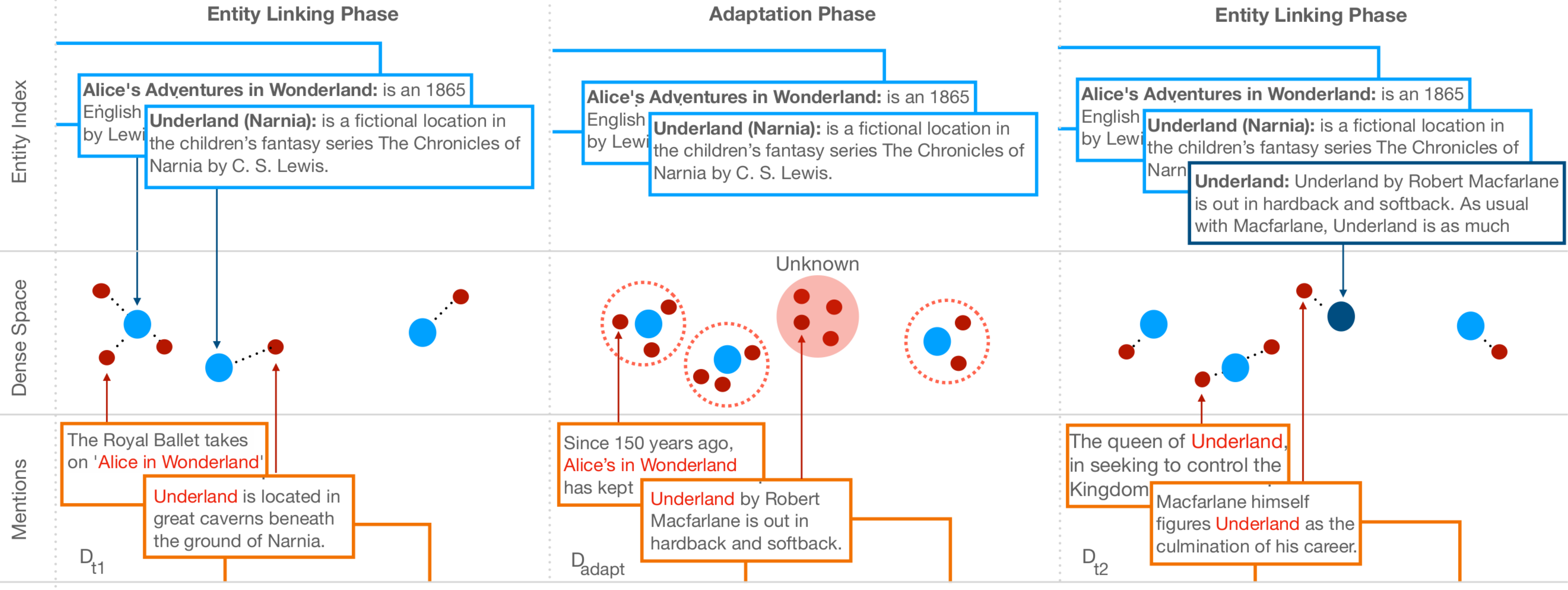}
\caption{\textbf{EDIN-pipeline:} In the adaptation phase, detected mentions in $D_{adapt}$ are mapped into a joint dense space with $E_{K}$ representations. A clustering algorithm groups mentions and entities based on kNN-similarity. Clusters of mentions without entity encoding are classified as $E_{U}$. To integrate these into the index of $E_{K}$, mentions in single sentence contexts are concatenated and mapped to a single embedding using the entity encoder. After adaptation, the updated entity index is used for standard EL.}
\label{fig:method}
\end{figure*}

\section{Model}
Our UEDI-pipeline is built on top of \cite{bela}, an end-to-end extension of the dense-retrieval based model BLINK \cite{wu2019zero}. It is composed of a Mention Detection (MD), Entity Disambiguation (ED) and Rejection (R) head. MD detects entity mention spans $[i, j]$ in context relying on BERT \cite{devlin-etal-2019-bert}. ED links these mentions to $e \in E_{K}$. It relies on bi-encoder architecture running a k-nearest-neighbor (kNN) search between \textit{mention encoding} and candidate \textit{entity encodings} (the entity index). Mention encodings are pooled from BERT-encoded paragraph tokens $p_{1..n}$:
\begin{align*}
\textbf{m}_{i,j} = FFL(BERT([CLS] p_{1} \ldots p_{n} [SEP])_{i...j})
\end{align*}
Entities are represented using BLINK's \textit{frozen} entity encoder:
\begin{align*}
\textbf{e} = BERT_{[CLS]}([CLS] t(e) [SEP] d(e) [SEP])
\end{align*}

Mention-entity candidates are passed to R that controls precision-recall trade-off by thresholding a learned candidate score.

More information about architecture and training are detailed in Appendix \ref{app:model}.

\section{Unknown Entity Discovery and Indexing}
For $E_{U}$ without descriptions, canonical names and labeled mentions, $L_{t1}$ is unable to successfully link mentions of the union of entities $E_{U} \bigcup E_{K}$.

To this end, we introduce an end-to-end pipeline to encode $E_{U}$ into $L_{t1}$'s entity index. The pipeline is depicted in Figure \ref{fig:method}. This pipeline is fully unsupervised and only relies on the adaptation dataset $D_{adapt}$. It follows a two-step process: i) Discovery: The entity linking system needs to detect mentions of unknown entities and classify them as unknown and ii) Indexing: co-referring mentions of unknown entities need to be mapped to a single embedding compatible with the entity index. 

\subsection{Unknown Entity Discovery}
\label{disc}
First, $L_{t1}$ detects and encodes mentions part of $D_{adapt}$. The MD head is trained to detect mentions leveraging the context around them, and can therefore detect mentions of both $E_{K}$ and $E_{U}$. Encoded mentions $ \textbf{M} =\{ \textbf{m}_{1}, . . . , \textbf{m}_{|M|}\} $ are then input to a clustering algorithm that partitions $M$ into disjoint clusters $ C =\{ c_{1}, . . . , c_{|C|}\} $. We use \citet{logan-iv-etal-2021-benchmarking}'s greedy NN clustering algorithm where $\textbf{m}_{i}$ is clustered with all $\textbf{m}_{j} \in \textbf{M}$ such that $sim(\textbf{m}_{i}, \textbf{m}_{j}) > \delta$.

Next, entity encodings of $e \in E_{K}$ are assigned to these clusters if $sim(\textbf{e}_{i}, \textbf{m}_{j}) > \tau$ and $\textbf{e}_{i}$ is the nearest entity of any $m_{i} \in c_{i}$. As we find this approach to result in low precision, we add another condition, namely that $sim(\textbf{e}_{i}, \textbf{m}_{j}) > \tau$ must hold for at least 70\% of all $\textbf{m}_{j} \in c_{i}$. $\delta, \tau$ are tuned on $D_{adapt}$-dev. For more details see Appendix C. Following \citet{agarwal2021entity}, all clusters not containing any entity representation are deemed to refer to entities in $E_{U}$. We refer to this subset of automatically identified unknown entities as $E'_{U}$.

\subsection{Unknown Entity Indexing}
Next, clusters identified as $E'_{U}$ are integrated into the EL index of $L_{t1}$. We explore two different methods of entity indexing: 
\begin{itemize}
\item \textit{Cluster-based} indexing: We concatenate all mentions part of the same cluster, each with the sentence they occur in, and use the entity encoder to map to a single entity representation. We pool over all $m_{i} \in c_{i}$ and select the most occurring mention as $t(e)$.
\item \textit{Mention-based} indexing: Mentions in single sentence contexts are indexed individually using the entity encoder.
Individual mentions are used as $t(e)$.

\end{itemize}

\section{Evaluation}
For \emph{Discovery}, we report precision and recall of $E_{U}$ classification and clustering metrics.

To compute standard EL metrics, e.g., precision and recall, canonical names of indexed clusters need to be consistent with the set of test labels. Our method of assigning canonical names to clusters based on mentions is not. To resolve this mismatch we pool on mention labels instead of the mentions itself. 

Unsupervised clustering of mentions in $D_{adapt}$ naturally comes with error: i) Clusters can be incomplete, e.g., mentions of a single entity can be split into multiple clusters which can lead to indexing the same entity multiple times and ii) Clusters can be impure, e.g., mentions of different entities end in the same cluster which leads to conflation of multiple entities into one representation. 

In our evaluation we make use of the gold labels for computing standard EL metrics by associating possibly more than one cluster to each unknown test entity, and considering a prediction correct if a test mention is linked to any of the clusters associated with the correct entity. Since, in a practical setting, gold labels are not available though, EL metrics could fail to capture shortcomings in establishing co-references between mentions.
Therefore, we report clustering metrics alongside EL metrics. We follow \citet{agarwal2021entity} and report normalized mutual information (NMI).

\section{Datasets}
\label{sec:Datasets}
\begin{table}[]
    \centering
    \footnotesize{
    \begin{tabular}{cl ccc }
    \toprule
      & \bf Wikipedia & \bf OSCAR \\
     \midrule
     Train &  100k (908k) & 100k (1.7M)  \\
     Adapt & 17k (183k) & 17k (360k) \\
     Dev Train & 8k (78k) & 8k (142k) \\
     Dev Adapt & - & 9k (183k)\\
     Test & 198k (1.8M)& 569k (11M) \\
    \bottomrule
    \end{tabular}
    \caption{\textbf{Dataset Statistics: } Number of samples (number of mentions) for training, adaptation and testing.}
    \label{tab:dataset_stats}}
\end{table}

\begin{table}[t]
\small
\centering
\begin{tabular}{lrrrr}
\toprule
\textbf{Bin} & Support & $E_{K}$ R@1 & Support & $E_{U}$ R@1 \\
\midrule
{[}0) & 68,241 & 21.1 & 7,095 & 17.5 \\
{[}1) & 59,227 & 29.1 & 3,923 & 25.9 \\
{[}1, 10) & 313,232 & 45.6 & 9,939 & 40.7 \\
{[}10, 100) & 901,857 & 65.7 & 7,765 & 57.3 \\
{[}100, 1k) & 2,860,880 & 76.9 & 7,399 & 64.4\\
{[}1k, +) & 5,981,028 & 84.4 & 6,717 & 86.7 \\
\bottomrule
\end{tabular}
\caption{\textbf{Frequency effects: }End-to-end EL performance of upper baseline model $L_{t2}$ per frequency bins.}
\label{tab:freq_bins_mentions}
\end{table}

\begin{figure}
\centering
\includegraphics[width=0.4\textwidth]{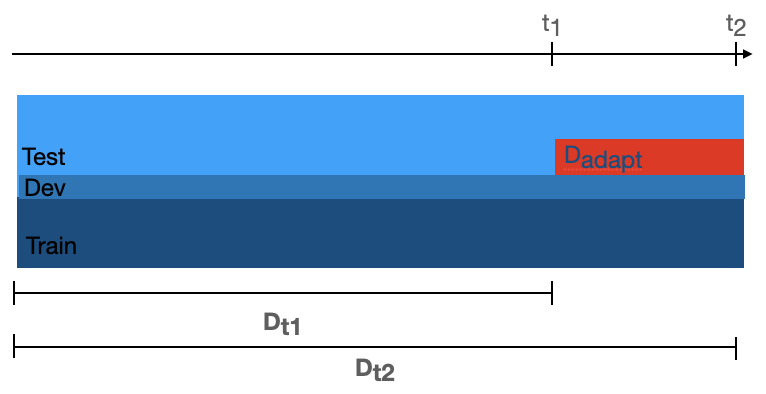}
\caption{\textbf{Dataset splits: } A schema illustrating the composition of $D_{t1}$ and $D_{t2}$. Note, that contrary to what this plot suggests, the number of samples per datasplit is equal for $D_{t1}$ and $D_{t2}$.}
\label{fig:datasplit}
\end{figure}

To construct the reference entity index, we download Wikipedia dumps from $t1$ and $t2$ and extract entity titles and descriptions. Setting $t_1$ to September 2019 (the date when BLINK was trained) the reference KB consists of 5.9M entities, setting $t_2$ to March 2022 an additional set of 0.7M entities is introduced.

Wikipedia and Oscar datasets are constructed as follows.
\\
\noindent
\textbf{Wikipedia:} Since usually only the first mention of an entity inside a Wikipedia article is hyperlinked, we annotate a subset of Wikipedia. We use a version of L that was trained at $t2$ on a labelled non-public dataset. While noisy, these predictions are significantly better than what our best discovery and indexing methods can achieve, therefore we adopt them as pseudo-labels for the purpose of comparing approaches. As discovery and indexing methods improve, manual labelling of the evaluation data will afford more accurate measures. Wikipedia provides time stamps which enables us to separate two time splits.\\
\\
\noindent
\textbf{OSCAR news:} This dataset is based on the common-crawl dataset OSCAR \cite{AbadjiOrtizSuarezRomaryetal.2021}. We select a subset of English language news pages which we label automatically as described above. The dataset consists of 797k samples, which we split based on their publication date.\\
\\

For both types of datasets we created two time splits: $D_1$, containing samples preceding $t1$, which is used to train model $L_{t1}$ and $D_2$, with samples preceding $t2$, which is used to train an upper bound model $L_{t2}$. To adapt $L_{t1}$, we hold out a subset of data from between $t1$ and $t2$ to construct $D_{adapt}$ ($D_{adapt} \cap D_2 = \emptyset$). Remaining samples are randomly split into train, dev, test. Figure \ref{fig:datasplit} illustrates the different data splits. Overall dataset statistics are listed in Table \ref{tab:dataset_stats}. 

To construct $D_{adapt}$, we follow \citet{agarwal2021entity}, and set a ratio of mentions of type $E_{U}$ to $E_{K}$ of 0.1. \footnote{Naturally this ratio would lie at 0.02. We made this artificial adjustment to reduce the strong class imbalance and obtain more interpretable and statistically stable results. Such adjustment could be lifted once considerably more precise unknown entity discovery components become available.
}
As $D_{t2}$-test covers both known and unknown entities, we use this dataset for EDIN-pipeline evaluation. In Oscar $D_{t2}$-test, the average number of mentions per $E_{U}$ is 5.6 and is ten times lower than for $E_{K}$. COVID-19 is the most occurring unknown entity with 12k mentions. 638k of $E_{U}$ are not mentioned at all and only 733 have a count larger 10. The distribution of number of $E_{U}$ is more skewed than the one of $E_{K}$ (skewness 529 vs. 106).

\section{Results and Discussion}

In the following sections, we report results for OSCAR data. Results on Wikipedia data are consistently lower than OSCAR results and shown in Appendix F. Our main findings are summarized in Table \ref{tab:e2e} (left) where we report end-to-end performance on OSCAR $D_{t2}$-test. As mentions of type $E_{U}$ are significantly less frequent than mentions of type $E_{K}$, we report results on these two types separately.

We first discuss upper ($L_{t2}$ a model trained on $D_{t2}$) and lower ($L_{t1}$ a model trained at $D_{t1}$) performance bounds. Next, we follow our two-step pipeline in adapting $L_{t1}$ such that it can successfully link mentions of $E_{U}$. Recall, that this pipeline involves two steps: discovery and indexing. Running $L_{t1}$ on $D_{adapt}$, we first discover mentions of $E_{U}$s by clustering. These clusters of mentions of $E_{U}$s are then indexed. We first present results on these components separately and then assemble the full end-to-end pipeline.

\begin{table*}[!htbp]
    \centering
    \begin{tabular}{lccc|ccc||ccc}
    \toprule
    &\multicolumn{3}{c}{\bf Known Entities} & \multicolumn{3}{c}{\bf Unknown Entities} & \multicolumn{3}{c}{\bf Unknown Entities filtered} \\
    \toprule
     \bf Model & \bf R@1 & \bf P@1 & \bf NMI & \bf R@1 & \bf P@1 & \bf NMI & \bf R@1 & \bf P@1 & \bf NMI \\
    \midrule[0.01pt]
    $L_{t1} $ & 80.1 & 82.0 & 93.5 & 0.0 & 0.0 & 0.0 & 0.0 & 0.0 & 0.0 \\
    $L_{t2}$  & 78.7 & 79.7 & 93.1 & 49.2 & 31.8 & 93.8 & 63.1 & 26.0 & 93.4\\
    \midrule
    $L_{t1}$-Descp  & 80.2 & 82.6 & 93.5 & 46.5 & 32.4 & 93.8 &58.3 & 26.2 & 90.5 \\
    \midrule[0.01pt]
    $L_{t1}$-Mention-Gold & 80.6 & 81.5 & 93.3 & 24.0 & 46.6 & 87.0 &40.7 & 46.6 & 87.0 \\
    $L_{t1}$-Mention & 80.3 & 81.9 & 93.4 & 20.5 & 43.7 & 87.6  & 34.5 & 43.5 & 88.7 \\
    \midrule[0.01pt]
    $L_{t1}$-Cluster-Gold & 80.3 & 82.0 & 94.2 & 30.5 & 51.8 & 85.9 & 51.8 & 51.8 & 85.9 \\
    EDIN ($L_{t1}$-Cluster) & 80.3 & 81.9 & 93.4 & 20.8 & 43.1 &  85.9 & 35.4 & 43.1 & 85.3 \\
    \bottomrule
    \end{tabular}
    \caption{\textbf{EL performance} on OSCAR $D_{t2}$-test for unknown entities $E_{U}$ and known entities $E_{K}$. \textbf{Left} shows end-to-end performance and \textbf{Right} shows filtered performance where mentions of $E_{U}$ which are not part of the oracle Cluster-based entity index are dropped from test. \textbf{Upper/Lower limit:} $L_{t1}$ uses Description-based entity representations, was trained at t1 and constitutes the lower performance bound. It lacks representations of $E_{U}$. $L_{t2}$ uses Description-based entity representations, was trained at t2 and constitutes the upper performance bound. $E_{U}$ are part of the entity index and labeled-mentions of $E_{U}$ were part of training. \textbf{Adaptation:} For $L_{t1}$-Descp Description-based entity representations are added to $L_{t1}$'s entity index. For $L_{t1}$-Mention Mention-based representations of i) oracle $E_{U}$ and ii) discovered $E'_{U}$ part of $D_{adapt}$ are added to $L_{t1}$'s entity index. For $L_{t1}$-Cluster Cluster-based representations of i) oracle $E_{U}$ and discovered $E'_{U}$ part of $D_{adapt}$ are added to $L_{t1}$'s entity index.}
    \label{tab:e2e}
\end{table*}

\subsection{Lower and upper bound}

Our starting point, and an obvious lower performance bound, is given by model $L_{t1}$.
This model lacks representations of $E_{U}$s and its training data does not contain any corresponding mentions. Therefore, performance on the subset of $E_{U}$s is 0 for all metrics.

For an upper performance bound we take model $L_{t2}$. The entities in $E_{U}$ were introduced to Wikipedia past $t_{1}$ but before $t_{2}$, meaning that to $L_{t2}$ these entities are actually \textit{known}: labeled mentions of $E_{U}$ are part of the training data and entity representations are part of the index. 

$L_{t2}$ reaches similar performance as $L_{t1}$ for $E_{K}$. We suspect performance differences can be attributed to the difference in training data.

Performance of $L_{t2}$ on mentions of $E_{U}$ is lower than on mentions of $E_{K}$. The performance discrepancy between $E_{U}$ and $E_{K}$ is largely due to frequency differences, see Table \ref{tab:freq_bins_mentions} where we report results per frequency band. We suspect that the remaining difference can be attributed to the degradation of the pre-trained LM and the entity encoder. Note that while labelled mentions of $E_{U}$ were seen during the training phase of L, BLINK's entity encoder was not retrained. To investigate this hypothesis further, we test $L_{t1}$ on mentions of $E_{K}$ that meet two conditions: i) time stamps of these samples are posterior to t1 and ii) two or more mentions of $E_{U}$ occur in their context. Thus, we target mentions of $E_{K}$ in novel contexts to which neither BLINK nor the PLM have been exposed. We find that recall drops only slightly from 80.1 to 79.9 but precision drops from 82.0 to 75.9. This result points to the conclusion that $E_{U}$ are also a source of noise when trying to link mentions of $E_{K}$.

\subsection{Discovery} 
\label{disc}
The first condition for effective discovery is the ability to reliably detect mentions of both known and unknown entities. Recall of $L_{t1}$ on $D_{adapt}$ for Mention Detection task is 90\% for $E_{K}$s and 86\% for $E_{U}$s entities. As expected, recall of mentions of $E_{K}$ is higher as no mentions of $E_{U}$ were seen during training. As a reference, running $L_{t2}$ on $D_{adapt}$, we find that for both $E_{K}$ and $E_{U}$ 91\% of mentions are recalled. Note again, that for $L_{t2}$, $E_{U}$ are \textit{known}. This indicates that mention detection is not affected by frequency differences and PLM degradation. 

Next, the clustering step to classify between mentions of $E_{U}$ and $E_{K}$ follows. We report clustering quality of 93.1\%  NMI.
When assigning entities to clusters as described in Section \ref{disc}, we find that our first attempt results in 35\% recall and 5\% precision on $D_{adapt}$-dev. This setting is close in spirit to \citet{agarwal2021entity} where a single entity-mention link is decisive for discovery.

A qualitative error analysis reveals that low recall is rooted in the problem that mention embeddings of $E_{U}$ (e.g. BioNTech) can have high similarity with entity representations of $E_{K}$ (e.g. of other biotechnology companies). If no mentions of these $E_{K}$ are part of $D_{adapt}$, we wrongly assign the high similarity entity to the cluster of mentions of $E_{U}$. We suspect that this problem is particularly pronounced in our setting as EDIN-benchmark is a large scale EL benchmark (up to 6 times more entities in the reference KB and up to 36 times more mentions in clustering set compared to \citet{agarwal2021entity}) with many long-tail entities (\citet{agarwal2021entity} entities were randomly selected). 

We find that low precision is rooted in the issue that $E_{K}$ are indexed as false positive when occurring in novel contexts, e.g., ``blood tests'' or ``vaccine'' in the context of COVID form their own distinct clusters.

With the modified cluster-entity assignment method, described in \ref{disc}, discovery still results in low but higher precision (8\%) and high recall (80\%) of $E_{U}$. 1196 of 1368 $E_{U}$ part of $D_{adapt}$ with more than three mentions are indexed.

\subsection{Indexing}

After discovery, we need to integrate clusters of mentions of $E_{U}$ into the existing entity index. In the following, we i) quantify the impact of re-training the model after indexing, ii) compare Description-based with Cluster-based indexing and iii) compare Cluster-based with Mention-based indexing. 

\subsubsection{Effect of re-training}

In the zero-shot EL problem, all entities are part of the index at training time. In the setting of EDIN-benchmark, discovery and indexing of $E_{U}$ happen after training. We run the following experiments to study the effect of this difference:

\begin{itemize}
    \item \textbf{Descp-post-train:} Description-based entity representations are added to the index \emph{after} training L on t1.
    \item \textbf{Descp-pre-train:} Description-based entity representations are added to the index \emph{before} training L on t1.
\end{itemize}

Recall and precision of $E_{U}$ in the setting where entity representations are added before the training is 47\% and 32\% respectively, see Table \ref{tab:descp}. %
Recall and precision in the setting where entity representations are added post training is 26\% and 17\% points lower respectively.

We note that entities in $E_{U}$ can potentially be placed in close proximity to $E_{K}$ in embedding space. When these entity encodings were present during training, they can be picked up as hard negatives and the mention encoder can learn to circumvent them. This hypothesis is supported by experiments showing that the mean similarity between mentions and correct known entity embeddings increases significantly when the mention encoder is re-trained after adding the new entities. For details see Appendix D.

The take-away for the EDIN-pipeline is that, after adding new entity representations to the index, another round of training is needed to adapt the mention encoder to the updated index. We adopt this approach for the following experiments. Besides adapting the mention encoder, retraining BLINK could have a similar effect: in such case learning from hard negative can affect the spacing of entity encodings. As re-training BLINK is expensive, we did not explore this option in this work.

\begin{table}[!htbp]
    \centering
    \resizebox{0.5\textwidth}{!}{\begin{tabular}{lccc|ccccc}
    \toprule
    & \multicolumn{3}{c}{\bf Unknown Entities} & \multicolumn{3}{c}{\bf Known Entities}\\
    \toprule
     & \bf R@1 & \bf P@1 & \bf NMI & \bf R@1 & \bf P@1 & \bf NMI \\
    \midrule[0.01pt]
    Not re-trained & 20.6 & 15.5 & 95.2 & 80.1 & 82.3 & 93.5 \\
    Re-trained & 46.5 & 32.4 & 93.8 & 80.2 & 82.6 & 93.5  \\
    \bottomrule
    \end{tabular}}
    \caption{\textbf{Effect of re-training:} End-to-end EL performance on OSCAR $D_{t2}$-test when adding Description-based representation of unknown entities $E_{U}$ to the entity index before and after training of $L_{t1}$.}
    \label{tab:descp}
\end{table}

\subsubsection{Description vs. Cluster-based indexing} 

We compared Description-based entity representations and Cluster-based indexing. To isolate discovery and indexing performance, we first evaluate performance using oracle clusters, where we replace the discovery method run on $D_{adapt}$ with an oracle where mentions of $E_{U}$ are discovered with perfect precision and recall and perfectly clustered. 

In this setting, Cluster-based indexing results in 19\% higher precision but 16\% lower recall, compared to Description-based indexing, see Table \ref{tab:e2e} (left) $L_{t1}$-Cluster-Gold. The drop in recall for Cluster-based indexing can partially be attributed to the fact that a subset of $E_{U}$ part of $D_{t2}$-test are not mentioned in $D_{adapt}$, and therefore no representation can be built for them. When filtering mentions of $E_{U}$ not part of $D_{adapt}$ out of the test set, the gap in recall decreases to 7\%, see Table \ref{tab:e2e} (right).

The take-away is that Cluster-based indexing relying on concatenated mentions in context instead of manually crafted descriptions has high precision potential but recall is challenging as a randomly selected adaptation set is unlikely to cover long-tail entities.

\subsubsection{Mention vs. Cluster-based indexing}

In this section, we compare Mention-based and Cluster-based indexing. Again, to isolate discovery and indexing we first report results using oracle clusters of $E_{U}$. When indexing mentions of these clusters individually, recall drops by 6\% points and precision by 5\%, see Table \ref{tab:e2e} (left), $L_{t1}$-Mention-Gold. When reducing the testset to mentions of entities that were actually discoverable, the difference in recall becomes even more pronounced: 40.7\% R@1 for Mention-based vs. 51.8 for Cluster-based indexing, see Table \ref{tab:e2e} (right). 

Interestingly, this means that the ability to attend over multiple mentions in context and unify their information into a single embedding leads to superior representations. Note that here the entity encoder was neither trained to deal with the style of individual mentions in context nor with clusters of mentions in context. For future work, it would be interesting to see if Cluster-based indexing can be generally beneficial to EL, outside of the context of EDIN-pipeline. This would require training the entity encoder to specifically adapt to the new input style.

\subsection{End-to-end model}

The prior section presented results of the two components individually. Now, we assemble the full end-to-end pipeline. We replace the oracle clusters of $E_{U}$ by discovered clusters of $E_{U}'$. Errors in discovery that affect indexing are: i) Clusters of mentions of $E_{K}$ classified false positively as unknown and clusters of mentions of $E_{U}$ false negatively classified as known and ii) incomplete and impure clusters. We find that performance of Mention-based and Cluster-based indexing in terms of R@1 and P@1 converges and is significantly lower than their oracle counterparts. Mention-based indexing has better clustering performance. 

When reducing the testset to mentions of entities that were actually discoverable, thus part of $D_{adapt}$, Cluster-based indexing is 1\% point better in terms of recall and 0.4 \% worse in precision, Table \ref{tab:e2e} (right). When reducing the testset further to mentions of entities that were in fact discovered, recall of Cluster-based indexing is with 58.4\% better than that of Mention-based indexing (55.5\%). 

Besides end-to-end performance, we also report entity disambiguation performance with oracle mention detection in Table \ref{tab:disambiguation} in Appendix E. Here, we find that Mention-based indexing is performing worse than Cluster-based indexing across all metrics: 1\% point difference in terms of recall, 3\% point difference in terms of precision and 0.7\% points difference in terms of NMI.

Overall, these results show that EDIN-benchmark is challenging. In the end-to-end setting, errors easily propagate. Most notably, we see this manifest when i) comparing Table \ref{tab:disambiguation} and \ref{tab:e2e} (right) results where the recall problem of $E_{U}$ becomes apparent and ii) comparing performance of oracle and automatic clusters where precision drops by 10\% points. Overall, we find that Cluster-based indexing, with the advantage of attending to and unifying the information of multiple mentions, performs better than Mention-based indexing. We call this version the EDIN-pipeline.
Besides yielding an index that scales in memory with the number of entities rather than the number of mentions -- a significant advantage when the number of entities is already large and in view of a streaming extension -- it generates fixed-size entity embeddings as a by-product that can have applications of their own and can be used to enhance PLMs (e.g., \citet{peters2019knowledge}).

In future work, we want to explore a setting where $E_{U}$ are discovered in a streaming fashion, thus scaling up $D_{adapt}$ and dropping the crafted 90\%-10\% ration of $E_{K}$ vs. $E_{U}$. This would pose challenges in terms of scale and precision in discovery. In terms of precision, a human in the loop approach, as proposed by \citet{Hoffart2016TheKA} in the context of keeping KBs fresh, to introduce a component of supervision into our end-to-end pipeline might be needed.

\section{Related work}
Entity linking is an extensively studied task. Prior to the introduction of PLMs, EL systems used frequency and typing information, alias tables, TF-IDF-based methods
and neural networks to model context, mention
and entity \cite{cucerzan-2007-large, bunescu-pasca-2006-using,Milne2008LearningTL, he-etal-2013-learning, Sun2015ModelingMC, lazic-etal-2015-plato, Raiman2018DeepTypeME, kolitsas-etal-2018-end, gupta-etal-2017-entity, ganea-hofmann-2017-deep, DBLP:journals/corr/abs-1811-10547, DBLP:journals/corr/abs-1909-05780}.

\citet{gillick-etal-2019-learning} present a  PLM-based dual encoder architecture that encodes mentions and entities in
the same dense vector space and performs EL via kNN search. \citet{logeswaran-etal-2019-zero} proposed the zero-shot EL task and show that domain adaptive training can address the domain shift problem. Subsequently, \citet{Wu2020ScalableZE} showed that pre-trained zero-shot architectures are both highly accurate and computationally efficient at scale. None of these works tackle the problem of unknown entities. 

Recently, \citet{fitzgerald-etal-2021-moleman} model EL entirely as mappings between mentions, where inference involves a NN search against all known mentions of all entities in the training set. In this setting mentions need to be labeled. They do not explore their approach in the setting of unknown entities.

Prior to dense retrieval-based EL, unknown entity discovery work includes:
\citet{ratinov-etal-2011-local} train a classifier to determine whether the top ranked EL candidate is unknown relying on local context, global Wikipedia coherence, and additional linker
features. Mentions of known entities that are incorrectly linked are among the training examples for unknown entity discovery thus decreasing quality.; \citet{nakashole-etal-2013-fine} introduce a model for unknown entity discovery and typing leveraging incompatibilities and correlations among entity
types. This method ignores context. Therefore, mentions not registered in the reference KB are regarded unknown; \citet{10.1145/2566486.2568003, Wu2016ExploringMF} study a variety of features for unknown entity discovery: \citet{10.1145/2566486.2568003} use perturbation-based confidence measures and key-phrase representations and \citet{Wu2016ExploringMF} explore different feature spaces, e.g., topical and search engine features. These features are not readily available and incorporating them into PLM-based approaches is not straightforward and outside the scope of this work.; \citet{DBLP:conf/tac/JiNHF15, derczynski-etal-2017-results} introduce shared tasks for discovery. These tasks do not target end-to-end dense-retrieval based EL and are outdated, e.g., treat iPhone as an unknown entity.; \citet{DBLP:conf/ijcai/Akasaki0T19} introduces a time sensitive method of discovering emerging entities. This method relies on training data in form of labeled emerging contexts. This data is collected by searching only exact mentions of unknown entities and therefore neglecting the issue of ambiguity and mention variety. 

None of these works consider unknown entities in an end-to-end setting including mention detection, unknown entity discovery and indexing. Also, we cannot use their datasets to evaluate as these entities were part of training the PLM.

Closely related to EL is the task of cross document entity co-reference (CDC), where no reference KB is present \cite{10.3115/980845.980859, gooi-allan-2004-cross, singh-etal-2011-large,dutta-weikum-2015-cross, barhom-etal-2019-revisiting, cattan-etal-2021-realistic, caciularu-etal-2021-cdlm-cross, cattan2021scico}. Most recently, \citet{logan-iv-etal-2021-benchmarking}
benchmark methods for streaming CDC, where mentions are disambiguated in a scalable manner via incremental clustering. Our work can be seen as bridging between the world of CDC and EL. We harvest CDC to discover and cluster unknown entities but then integrate them into a curated list of entities. \citet{dutta-weikum-2015-c3el}
also combine clustering-based CDC decisions and linking but as this work is using sparse bag-of-word representations, it is not well suited for the embedding-based representations used in this work.

Most recently, \citet{angell-etal-2021-clustering} introduce a new EL method using document-level supervised graph-based clustering. \citet{agarwal2021entity} extend this work to cross-document EL and entity discovery. In this work, we adopt a more standard bi-encoder architecture (i.e. BLINK), with better EL scalability potential (memory linear in the number of entities and not in the number of mentions) and an existing end-to-end extension. We use a modified version of their discovery method.

Besides dense-retrieval based EL, \citet{decao2020autoregressive} proposed generative EL. We plan to study the problem of integrating unknown entities into generative models in future work.

\section{Conclusion}
This work created EDIN benchmark and pipeline. EDIN-benchmark is a large-scale, end-to-end entity linking benchmark with a clear cut temporal segmentation for \emph{Unknown Entity Discovery and Indexing}. EDIN-pipeline detects and clusters mentions of unknown entities in context. These clusters of unknown mentions are then collapsed into single embeddings and integrated into the entity index of the original entity linking system.
\bibliographystyle{acl_natbib}
\bibliography{bibliography,acl2021}

\begin{thebibliography}{49}
\expandafter\ifx\csname natexlab\endcsname\relax\def\natexlab#1{#1}\fi

\bibitem[{Abadji et~al.(2021)Abadji, Su{\'a}rez, Romary, and
  Sagot}]{AbadjiOrtizSuarezRomaryetal.2021}
Julien Abadji, Pedro Javier~Ortiz Su{\'a}rez, Laurent Romary, and Beno{\^i}t
  Sagot. 2021.
\newblock \href {https://doi.org/10.14618/ids-pub-10468} {Ungoliant: An
  optimized pipeline for the generation of a very large-scale multilingual web
  corpus}.
\newblock Proceedings of the Workshop on Challenges in the Management of Large
  Corpora (CMLC-9) 2021. Limerick, 12 July 2021 (Online-Event), pages 1 -- 9,
  Mannheim. Leibniz-Institut f{\"u}r Deutsche Sprache.

\bibitem[{Agarwal et~al.(2021)Agarwal, Angell, Monath, and
  McCallum}]{agarwal2021entity}
Dhruv Agarwal, Rico Angell, Nicholas Monath, and Andrew McCallum. 2021.
\newblock \href {http://arxiv.org/abs/2109.01242} {Entity linking and discovery
  via arborescence-based supervised clustering}.

\bibitem[{Akasaki et~al.(2019)Akasaki, Yoshinaga, and
  Toyoda}]{DBLP:conf/ijcai/Akasaki0T19}
Satoshi Akasaki, Naoki Yoshinaga, and Masashi Toyoda. 2019.
\newblock \href {https://doi.org/10.24963/ijcai.2019/678} {Early discovery of
  emerging entities in microblogs}.
\newblock In \emph{Proceedings of the Twenty-Eighth International Joint
  Conference on Artificial Intelligence, {IJCAI} 2019, Macao, China, August
  10-16, 2019}, pages 4882--4889. ijcai.org.

\bibitem[{Angell et~al.(2021)Angell, Monath, Mohan, Yadav, and
  McCallum}]{angell-etal-2021-clustering}
Rico Angell, Nicholas Monath, Sunil Mohan, Nishant Yadav, and Andrew McCallum.
  2021.
\newblock \href {https://doi.org/10.18653/v1/2021.naacl-main.205}
  {Clustering-based inference for biomedical entity linking}.
\newblock In \emph{Proceedings of the 2021 Conference of the North American
  Chapter of the Association for Computational Linguistics: Human Language
  Technologies}, pages 2598--2608, Online. Association for Computational
  Linguistics.

\bibitem[{Anonymous(2022)}]{bela}
Anonymous. 2022.
\newblock End to end entity linking in 100 languages.

\bibitem[{Bagga and Baldwin(1998)}]{10.3115/980845.980859}
Amit Bagga and Breck Baldwin. 1998.
\newblock \href {https://doi.org/10.3115/980845.980859} {Entity-based
  cross-document coreferencing using the vector space model}.
\newblock In \emph{Proceedings of the 36th Annual Meeting of the Association
  for Computational Linguistics and 17th International Conference on
  Computational Linguistics - Volume 1}, ACL '98/COLING '98, page 79–85, USA.
  Association for Computational Linguistics.

\bibitem[{Barhom et~al.(2019)Barhom, Shwartz, Eirew, Bugert, Reimers, and
  Dagan}]{barhom-etal-2019-revisiting}
Shany Barhom, Vered Shwartz, Alon Eirew, Michael Bugert, Nils Reimers, and Ido
  Dagan. 2019.
\newblock \href {https://doi.org/10.18653/v1/P19-1409} {Revisiting joint
  modeling of cross-document entity and event coreference resolution}.
\newblock In \emph{Proceedings of the 57th Annual Meeting of the Association
  for Computational Linguistics}, pages 4179--4189, Florence, Italy.
  Association for Computational Linguistics.

\bibitem[{Bunescu and Pa{\c{s}}ca(2006)}]{bunescu-pasca-2006-using}
Razvan Bunescu and Marius Pa{\c{s}}ca. 2006.
\newblock \href {https://aclanthology.org/E06-1002} {Using encyclopedic
  knowledge for named entity disambiguation}.
\newblock In \emph{11th Conference of the {E}uropean Chapter of the Association
  for Computational Linguistics}, pages 9--16, Trento, Italy. Association for
  Computational Linguistics.

\bibitem[{Caciularu et~al.(2021)Caciularu, Cohan, Beltagy, Peters, Cattan, and
  Dagan}]{caciularu-etal-2021-cdlm-cross}
Avi Caciularu, Arman Cohan, Iz~Beltagy, Matthew Peters, Arie Cattan, and Ido
  Dagan. 2021.
\newblock \href {https://doi.org/10.18653/v1/2021.findings-emnlp.225} {{CDLM}:
  Cross-document language modeling}.
\newblock In \emph{Findings of the Association for Computational Linguistics:
  EMNLP 2021}, pages 2648--2662, Punta Cana, Dominican Republic. Association
  for Computational Linguistics.

\bibitem[{Cao et~al.(2018)Cao, Hou, Li, and Liu}]{Cao2018NeuralCE}
Yixin Cao, Lei Hou, Juan-Zi Li, and Zhiyuan Liu. 2018.
\newblock Neural collective entity linking.
\newblock In \emph{COLING}.

\bibitem[{Cattan et~al.(2021{\natexlab{a}})Cattan, Eirew, Stanovsky, Joshi, and
  Dagan}]{cattan-etal-2021-realistic}
Arie Cattan, Alon Eirew, Gabriel Stanovsky, Mandar Joshi, and Ido Dagan.
  2021{\natexlab{a}}.
\newblock \href {https://doi.org/10.18653/v1/2021.starsem-1.13} {Realistic
  evaluation principles for cross-document coreference resolution}.
\newblock In \emph{Proceedings of *SEM 2021: The Tenth Joint Conference on
  Lexical and Computational Semantics}, pages 143--151, Online. Association for
  Computational Linguistics.

\bibitem[{Cattan et~al.(2021{\natexlab{b}})Cattan, Johnson, Weld, Dagan,
  Beltagy, Downey, and Hope}]{cattan2021scico}
Arie Cattan, Sophie Johnson, Daniel~S. Weld, Ido Dagan, Iz~Beltagy, Doug
  Downey, and Tom Hope. 2021{\natexlab{b}}.
\newblock \href {https://openreview.net/forum?id=OFLbgUP04nC} {Scico:
  Hierarchical cross-document coreference for scientific concepts}.
\newblock In \emph{3rd Conference on Automated Knowledge Base Construction}.

\bibitem[{Cucerzan(2007)}]{cucerzan-2007-large}
Silviu Cucerzan. 2007.
\newblock \href {https://aclanthology.org/D07-1074} {Large-scale named entity
  disambiguation based on {W}ikipedia data}.
\newblock In \emph{Proceedings of the 2007 Joint Conference on Empirical
  Methods in Natural Language Processing and Computational Natural Language
  Learning ({EMNLP}-{C}o{NLL})}, pages 708--716, Prague, Czech Republic.
  Association for Computational Linguistics.

\bibitem[{De~Cao et~al.(2021)De~Cao, Izacard, Riedel, and
  Petroni}]{decao2020autoregressive}
Nicola De~Cao, Gautier Izacard, Sebastian Riedel, and Fabio Petroni. 2021.
\newblock \href {https://openreview.net/forum?id=5k8F6UU39V} {Autoregressive
  entity retrieval}.
\newblock In \emph{International Conference on Learning Representations}.

\bibitem[{Derczynski et~al.(2017)Derczynski, Nichols, van Erp, and
  Limsopatham}]{derczynski-etal-2017-results}
Leon Derczynski, Eric Nichols, Marieke van Erp, and Nut Limsopatham. 2017.
\newblock \href {https://doi.org/10.18653/v1/W17-4418} {Results of the
  {WNUT}2017 shared task on novel and emerging entity recognition}.
\newblock In \emph{Proceedings of the 3rd Workshop on Noisy User-generated
  Text}, pages 140--147, Copenhagen, Denmark. Association for Computational
  Linguistics.

\bibitem[{Devlin et~al.(2019)Devlin, Chang, Lee, and
  Toutanova}]{devlin-etal-2019-bert}
Jacob Devlin, Ming-Wei Chang, Kenton Lee, and Kristina Toutanova. 2019.
\newblock \href {https://doi.org/10.18653/v1/N19-1423} {{BERT}: Pre-training of
  deep bidirectional transformers for language understanding}.
\newblock In \emph{Proceedings of the 2019 Conference of the North {A}merican
  Chapter of the Association for Computational Linguistics: Human Language
  Technologies, Volume 1 (Long and Short Papers)}, pages 4171--4186,
  Minneapolis, Minnesota. Association for Computational Linguistics.

\bibitem[{Dhingra et~al.(2022)Dhingra, Cole, Eisenschlos, Gillick, Eisenstein,
  and Cohen}]{TACL3539}
Bhuwan Dhingra, Jeremy Cole, Julian Eisenschlos, Daniel Gillick, Jacob
  Eisenstein, and William Cohen. 2022.
\newblock \href {https://transacl.org/index.php/tacl/article/view/3539}
  {Time-aware language models as temporal knowledge bases}.
\newblock \emph{Transactions of the Association for Computational Linguistics},
  10(0):257--273.

\bibitem[{Dutta and Weikum(2015{\natexlab{a}})}]{dutta-weikum-2015-c3el}
Sourav Dutta and Gerhard Weikum. 2015{\natexlab{a}}.
\newblock \href {https://doi.org/10.18653/v1/D15-1101} {{C}3{EL}: A joint model
  for cross-document co-reference resolution and entity linking}.
\newblock In \emph{Proceedings of the 2015 Conference on Empirical Methods in
  Natural Language Processing}, pages 846--856, Lisbon, Portugal. Association
  for Computational Linguistics.

\bibitem[{Dutta and Weikum(2015{\natexlab{b}})}]{dutta-weikum-2015-cross}
Sourav Dutta and Gerhard Weikum. 2015{\natexlab{b}}.
\newblock \href {https://doi.org/10.1162/tacl_a_00119} {Cross-document
  co-reference resolution using sample-based clustering with knowledge
  enrichment}.
\newblock \emph{Transactions of the Association for Computational Linguistics},
  3:15--28.

\bibitem[{FitzGerald et~al.(2021)FitzGerald, Bikel, Botha, Gillick,
  Kwiatkowski, and McCallum}]{fitzgerald-etal-2021-moleman}
Nicholas FitzGerald, Dan Bikel, Jan Botha, Daniel Gillick, Tom Kwiatkowski, and
  Andrew McCallum. 2021.
\newblock \href {https://doi.org/10.18653/v1/2021.acl-short.37} {{MOLEMAN}:
  Mention-only linking of entities with a mention annotation network}.
\newblock In \emph{Proceedings of the 59th Annual Meeting of the Association
  for Computational Linguistics and the 11th International Joint Conference on
  Natural Language Processing (Volume 2: Short Papers)}, pages 278--285,
  Online. Association for Computational Linguistics.

\bibitem[{Ganea and Hofmann(2017)}]{ganea-hofmann-2017-deep}
Octavian-Eugen Ganea and Thomas Hofmann. 2017.
\newblock \href {https://doi.org/10.18653/v1/D17-1277} {Deep joint entity
  disambiguation with local neural attention}.
\newblock In \emph{Proceedings of the 2017 Conference on Empirical Methods in
  Natural Language Processing}, pages 2619--2629, Copenhagen, Denmark.
  Association for Computational Linguistics.

\bibitem[{Gillick et~al.(2019)Gillick, Kulkarni, Lansing, Presta, Baldridge,
  Ie, and Garcia-Olano}]{gillick-etal-2019-learning}
Daniel Gillick, Sayali Kulkarni, Larry Lansing, Alessandro Presta, Jason
  Baldridge, Eugene Ie, and Diego Garcia-Olano. 2019.
\newblock \href {https://doi.org/10.18653/v1/K19-1049} {Learning dense
  representations for entity retrieval}.
\newblock In \emph{Proceedings of the 23rd Conference on Computational Natural
  Language Learning (CoNLL)}, pages 528--537, Hong Kong, China. Association for
  Computational Linguistics.

\bibitem[{Gooi and Allan(2004)}]{gooi-allan-2004-cross}
Chung~Heong Gooi and James Allan. 2004.
\newblock \href {https://aclanthology.org/N04-1002} {Cross-document coreference
  on a large scale corpus}.
\newblock In \emph{Proceedings of the Human Language Technology Conference of
  the North {A}merican Chapter of the Association for Computational
  Linguistics: {HLT}-{NAACL} 2004}, pages 9--16, Boston, Massachusetts, USA.
  Association for Computational Linguistics.

\bibitem[{Gupta et~al.(2017)Gupta, Singh, and Roth}]{gupta-etal-2017-entity}
Nitish Gupta, Sameer Singh, and Dan Roth. 2017.
\newblock \href {https://doi.org/10.18653/v1/D17-1284} {Entity linking via
  joint encoding of types, descriptions, and context}.
\newblock In \emph{Proceedings of the 2017 Conference on Empirical Methods in
  Natural Language Processing}, pages 2681--2690, Copenhagen, Denmark.
  Association for Computational Linguistics.

\bibitem[{He et~al.(2013)He, Liu, Li, Zhou, Zhang, and
  Wang}]{he-etal-2013-learning}
Zhengyan He, Shujie Liu, Mu~Li, Ming Zhou, Longkai Zhang, and Houfeng Wang.
  2013.
\newblock \href {https://aclanthology.org/P13-2006} {Learning entity
  representation for entity disambiguation}.
\newblock In \emph{Proceedings of the 51st Annual Meeting of the Association
  for Computational Linguistics (Volume 2: Short Papers)}, pages 30--34, Sofia,
  Bulgaria. Association for Computational Linguistics.

\bibitem[{Hoffart et~al.(2014)Hoffart, Altun, and
  Weikum}]{10.1145/2566486.2568003}
Johannes Hoffart, Yasemin Altun, and Gerhard Weikum. 2014.
\newblock \href {https://doi.org/10.1145/2566486.2568003} {Discovering emerging
  entities with ambiguous names}.
\newblock In \emph{Proceedings of the 23rd International Conference on World
  Wide Web}, WWW '14, page 385–396, New York, NY, USA. Association for
  Computing Machinery.

\bibitem[{Hoffart et~al.(2016)Hoffart, Milchevski, Weikum, Anand, and
  Singh}]{Hoffart2016TheKA}
Johannes Hoffart, Dragan Milchevski, Gerhard Weikum, Avishek Anand, and
  Jaspreet Singh. 2016.
\newblock The knowledge awakens: Keeping knowledge bases fresh with emerging
  entities.
\newblock \emph{Proceedings of the 25th International Conference Companion on
  World Wide Web}.

\bibitem[{Ji et~al.(2015)Ji, Nothman, Hachey, and
  Florian}]{DBLP:conf/tac/JiNHF15}
Heng Ji, Joel Nothman, Ben Hachey, and Radu Florian. 2015.
\newblock \href
  {https://tac.nist.gov/publications/2015/additional.papers/TAC2015.KBP\_Trilingual\_Entity\_Discovery\_and\_Linking\_overview.proceedings.pdf}
  {Overview of {TAC-KBP2015} tri-lingual entity discovery and linking}.
\newblock In \emph{Proceedings of the 2015 Text Analysis Conference, {TAC}
  2015, Gaithersburg, Maryland, USA, November 16-17, 2015, 2015}. {NIST}.

\bibitem[{Johnson et~al.(2017)Johnson, Douze, and J{\'e}gou}]{JDH17}
Jeff Johnson, Matthijs Douze, and Herv{\'e} J{\'e}gou. 2017.
\newblock Billion-scale similarity search with gpus.
\newblock \emph{arXiv preprint arXiv:1702.08734}.

\bibitem[{Khalife and Vazirgiannis(2018)}]{DBLP:journals/corr/abs-1811-10547}
Sammy Khalife and Michalis Vazirgiannis. 2018.
\newblock \href {http://arxiv.org/abs/1811.10547} {Scalable graph-based
  individual named entity identification}.
\newblock \emph{CoRR}, abs/1811.10547.

\bibitem[{Kolitsas et~al.(2018)Kolitsas, Ganea, and
  Hofmann}]{kolitsas-etal-2018-end}
Nikolaos Kolitsas, Octavian-Eugen Ganea, and Thomas Hofmann. 2018.
\newblock \href {https://doi.org/10.18653/v1/K18-1050} {End-to-end neural
  entity linking}.
\newblock In \emph{Proceedings of the 22nd Conference on Computational Natural
  Language Learning}, pages 519--529, Brussels, Belgium. Association for
  Computational Linguistics.

\bibitem[{Lazaridou et~al.(2021)Lazaridou, Kuncoro, Gribovskaya, Agrawal,
  Liska, Terzi, Gimenez, de~Masson~d\textquotesingle Autume, Kocisky, Ruder,
  Yogatama, Cao, Young, and Blunsom}]{NEURIPS2021_f5bf0ba0}
Angeliki Lazaridou, Adhi Kuncoro, Elena Gribovskaya, Devang Agrawal, Adam
  Liska, Tayfun Terzi, Mai Gimenez, Cyprien de~Masson~d\textquotesingle Autume,
  Tomas Kocisky, Sebastian Ruder, Dani Yogatama, Kris Cao, Susannah Young, and
  Phil Blunsom. 2021.
\newblock \href
  {https://proceedings.neurips.cc/paper/2021/file/f5bf0ba0a17ef18f9607774722f5698c-Paper.pdf}
  {Mind the gap: Assessing temporal generalization in neural language models}.
\newblock In \emph{Advances in Neural Information Processing Systems},
  volume~34, pages 29348--29363. Curran Associates, Inc.

\bibitem[{Lazic et~al.(2015)Lazic, Subramanya, Ringgaard, and
  Pereira}]{lazic-etal-2015-plato}
Nevena Lazic, Amarnag Subramanya, Michael Ringgaard, and Fernando Pereira.
  2015.
\newblock \href {https://doi.org/10.1162/tacl_a_00154} {{P}lato: A selective
  context model for entity resolution}.
\newblock \emph{Transactions of the Association for Computational Linguistics},
  3:503--515.

\bibitem[{Ledell~Wu(2020)}]{wu2019zero}
Martin Josifoski Sebastian Riedel Luke~Zettlemoyer Ledell~Wu, Fabio~Petroni.
  2020.
\newblock Zero-shot entity linking with dense entity retrieval.
\newblock In \emph{EMNLP}.

\bibitem[{Li et~al.(2020)Li, Min, Iyer, Mehdad, and Yih}]{li2020efficient}
Belinda~Z. Li, Sewon Min, Srinivasan Iyer, Yashar Mehdad, and Wen-tau Yih.
  2020.
\newblock Efficient one-pass end-to-end entity linking for questions.
\newblock In \emph{EMNLP}.

\bibitem[{Logan~IV et~al.(2021)Logan~IV, McCallum, Singh, and
  Bikel}]{logan-iv-etal-2021-benchmarking}
Robert~L Logan~IV, Andrew McCallum, Sameer Singh, and Dan Bikel. 2021.
\newblock \href {https://doi.org/10.18653/v1/2021.acl-long.364} {Benchmarking
  scalable methods for streaming cross document entity coreference}.
\newblock In \emph{Proceedings of the 59th Annual Meeting of the Association
  for Computational Linguistics and the 11th International Joint Conference on
  Natural Language Processing (Volume 1: Long Papers)}, pages 4717--4731,
  Online. Association for Computational Linguistics.

\bibitem[{Logeswaran et~al.(2019)Logeswaran, Chang, Lee, Toutanova, Devlin, and
  Lee}]{logeswaran-etal-2019-zero}
Lajanugen Logeswaran, Ming-Wei Chang, Kenton Lee, Kristina Toutanova, Jacob
  Devlin, and Honglak Lee. 2019.
\newblock \href {https://doi.org/10.18653/v1/P19-1335} {Zero-shot entity
  linking by reading entity descriptions}.
\newblock In \emph{Proceedings of the 57th Annual Meeting of the Association
  for Computational Linguistics}, pages 3449--3460, Florence, Italy.
  Association for Computational Linguistics.

\bibitem[{Milne and Witten(2008)}]{Milne2008LearningTL}
David~N. Milne and Ian~H. Witten. 2008.
\newblock Learning to link with wikipedia.
\newblock In \emph{CIKM '08}.

\bibitem[{Nakashole et~al.(2013)Nakashole, Tylenda, and
  Weikum}]{nakashole-etal-2013-fine}
Ndapandula Nakashole, Tomasz Tylenda, and Gerhard Weikum. 2013.
\newblock \href {https://aclanthology.org/P13-1146} {Fine-grained semantic
  typing of emerging entities}.
\newblock In \emph{Proceedings of the 51st Annual Meeting of the Association
  for Computational Linguistics (Volume 1: Long Papers)}, pages 1488--1497,
  Sofia, Bulgaria. Association for Computational Linguistics.

\bibitem[{Onoe and Durrett(2019)}]{DBLP:journals/corr/abs-1909-05780}
Yasumasa Onoe and Greg Durrett. 2019.
\newblock \href {http://arxiv.org/abs/1909.05780} {Fine-grained entity typing
  for domain independent entity linking}.
\newblock \emph{CoRR}, abs/1909.05780.

\bibitem[{Onoe and Durrett(2020)}]{Onoe2020FineGrainedET}
Yasumasa Onoe and Greg Durrett. 2020.
\newblock Fine-grained entity typing for domain independent entity linking.
\newblock In \emph{AAAI}.

\bibitem[{Peters et~al.(2019)Peters, Neumann, Logan, Schwartz, Joshi, Singh,
  and Smith}]{peters2019knowledge}
Matthew~E Peters, Mark Neumann, Robert Logan, Roy Schwartz, Vidur Joshi, Sameer
  Singh, and Noah~A Smith. 2019.
\newblock Knowledge enhanced contextual word representations.
\newblock In \emph{Conference on Empirical Methods in Natural Language
  Processing and the 9th International Joint Conference on Natural Language
  Processing (EMNLP-IJCNLP)}.

\bibitem[{Raiman and Raiman(2018)}]{Raiman2018DeepTypeME}
Jonathan Raiman and Olivier Raiman. 2018.
\newblock Deeptype: Multilingual entity linking by neural type system
  evolution.
\newblock In \emph{AAAI}.

\bibitem[{Ratinov et~al.(2011)Ratinov, Roth, Downey, and
  Anderson}]{ratinov-etal-2011-local}
Lev Ratinov, Dan Roth, Doug Downey, and Mike Anderson. 2011.
\newblock \href {https://aclanthology.org/P11-1138} {Local and global
  algorithms for disambiguation to {W}ikipedia}.
\newblock In \emph{Proceedings of the 49th Annual Meeting of the Association
  for Computational Linguistics: Human Language Technologies}, pages
  1375--1384, Portland, Oregon, USA. Association for Computational Linguistics.

\bibitem[{Singh et~al.(2011)Singh, Subramanya, Pereira, and
  McCallum}]{singh-etal-2011-large}
Sameer Singh, Amarnag Subramanya, Fernando Pereira, and Andrew McCallum. 2011.
\newblock \href {https://aclanthology.org/P11-1080} {Large-scale cross-document
  coreference using distributed inference and hierarchical models}.
\newblock In \emph{Proceedings of the 49th Annual Meeting of the Association
  for Computational Linguistics: Human Language Technologies}, pages 793--803,
  Portland, Oregon, USA. Association for Computational Linguistics.

\bibitem[{Sun et~al.(2015{\natexlab{a}})Sun, Lin, Tang, Yang, Ji, and
  Wang}]{Sun2015ModelingMC}
Yaming Sun, Lei Lin, Duyu Tang, Nan Yang, Zhenzhou Ji, and Xiaolong Wang.
  2015{\natexlab{a}}.
\newblock Modeling mention, context and entity with neural networks for entity
  disambiguation.
\newblock In \emph{IJCAI}.

\bibitem[{Sun et~al.(2015{\natexlab{b}})Sun, Lin, Tang, Yang, Ji, and
  Wang}]{10.5555/2832415.2832435}
Yaming Sun, Lei Lin, Duyu Tang, Nan Yang, Zhenzhou Ji, and Xiaolong Wang.
  2015{\natexlab{b}}.
\newblock Modeling mention, context and entity with neural networks for entity
  disambiguation.
\newblock In \emph{Proceedings of the 24th International Conference on
  Artificial Intelligence}, IJCAI'15, page 1333–1339. AAAI Press.

\bibitem[{Wu et~al.(2020)Wu, Petroni, Josifoski, Riedel, and
  Zettlemoyer}]{Wu2020ScalableZE}
Ledell~Yu Wu, Fabio Petroni, Martin Josifoski, Sebastian Riedel, and Luke
  Zettlemoyer. 2020.
\newblock Scalable zero-shot entity linking with dense entity retrieval.
\newblock In \emph{EMNLP}.

\bibitem[{Wu et~al.(2016)Wu, Song, and Giles}]{Wu2016ExploringMF}
Zhaohui Wu, Yang Song, and C.~Lee Giles. 2016.
\newblock Exploring multiple feature spaces for novel entity discovery.
\newblock In \emph{AAAI}.

\end{thebibliography}

\appendix
\section{Model}
\label{app:model}
In the following sections, we explain \cite{bela}'s architecture in detail.

\subsection{Mention Detection}

For every span [i,j], the MD head calculates the probability of [i,j] being the mention of an entity by scoring whether i is the start of the mention, j is the end of the mention, and the tokens between i and j are the insides:

\begin{align*}
& s_{start(i)} = \textbf{w}_{start}^{T}  \textbf{p}_{i} \\
& s_{end(i)} = \textbf{w}_{end}^{T}  \textbf{p}_{j} \\
& s_{mention(t)} = \textbf{w}_{mention}^{T}  \textbf{p}_{t}
\end{align*}
where $\textbf{w}_{start}, \textbf{w}_{end}, \textbf{w}_{mention}$ are learnable vectors and $\textbf{p}_{i}$ paragraph token representations based on BERT:
\begin{align*}
[\textbf{p}_{1} \ldots \textbf{p}_{n}] = BERT([CLS] p_{1} \ldots p_{n} [SEP])
\end{align*}
Overall mention probabilities are computed as:
\begin{align*}
p([i, j]) = \sigma(s_{start(i)} + s_{end(j)} + \sum_{t=i}^{j} (s_{mention(t)}))
\end{align*}
Top candidates are selected as mention candidates and propagate to the next step.

\subsection{Entity Disambiguation}
The ED head receives mention spans in the text and finds the best matching entity in the KB.

Following \citet{Wu2020ScalableZE}, ED is based on dense retrieval. Description-based entity representations are computed as follows:

\begin{align*}
\textbf{e} = BERT_{[CLS]}([CLS] t(e) [SEP] d(e) [SEP])
\end{align*}
Following \citet{li2020efficient}, mention representations are constructed with one pass of the encoder and without mention boundary tokens by pooling mention tokens from the encoder output:
\begin{align*}
\textbf{m}_{i,j} = FFL(\textbf{p}_{i} \ldots  \textbf{p}_{j})
\end{align*}

Similarity score $s$ between the mention candidate and an entity candidate $e \in E$ are computed:
\begin{align*}
s(e, [i, j]) = \textbf{e}*\textbf{m}_{i,j}
\end{align*}

A likelihood distribution over
all entities, conditioned on the mention $[i, j]$ is computed:
\begin{align*}
p(e|[i, j]) = \frac{exp(s(e, [i, j]))}{\sum_{e' \in E} exp(s(e', [i, j]))}
\end{align*}
$<[i,j],e^*>$, such that \begin{align*}e^* = argmax _{e}(p([i,j],e)),\end{align*} are passed as a candidate $<$ mention span, entity $>$ tuple to the rejection head.

\subsection{Rejection head}
MD and ED steps over-generate. R looks at an $(e^*, [i, j])$ pair holistically decides whether to accept it. Input features to R are the MD score $p([i, j])$, the ED score $p(e^*|[i, j])$, the mention representation $\textbf{y}_{i,j}$, top-ranked candidate representation $\textbf{x}_{e^*}$ as well as their difference and Hadamard product. The concatenation of these features is fed through a feed-forward network to output the final entity linking score $p([i, j], e^*)$. All $p([i, j], e^*)>\gamma$ are accepted where $\gamma$ is a threshold set to 0.4.

\subsection{Training}
Following prior work \cite{10.5555/2832415.2832435, Cao2018NeuralCE, gillick-etal-2019-learning, Onoe2020FineGrainedET}, training is split into two stages. First, ED only is trained on a Wikipedia dataset. This dataset is constructed by extracting Wikipedia hyperlinks to labeled mention-entity pairs and consists of 17M training samples. Then, ED, MD and R are trained jointly on the downstream dataset (either Oscar or Wikipedia).
To train the ED head, frozen entity representations are used. As entity embeddings do not change during training, entity embeddings can be indexed using quantization algorithms for a fast kNN search (using FAISS \cite{JDH17} framework with HNSW index). A likelihood distribution over positive and mined hard negative entities for each mention is computed. Negative Log-Likelihood loss across all gold mentions in the text is used.

To train MD, binary cross-entropy loss between all possible valid spans and gold mentions in the training set is computed.

To train R, binary cross-entropy loss between retrieved mention-entity pairs and gold mention-entity pairs is used.

\section{OSCAR-based dataset}
\label{sec:oscar}
We select the following six online news pages:\\
\\
\noindent
BBC: https://www.bbc.com/\\
CNN: https://www.cnn.com/\\
Deutsche Welle: https://www.dw.com/en/\\
Reuters: https://www.reuters.com/article/\\
Guardian: https://www.theguardian.com/\\ Associated Press: https://apnews.com/article/

\section{Hyper-parameters Adaptation phase}
\label{sec:clustering}
Using OSCAR $D_{adapt}$-dev, we optimize mention score threshold $s_{m}$, greedy NN distance threshold $d_{m}$ and mention entity similarity threshold $d_{e}$. 

We optimize $s_{m}$ in range 0.0 to 1.0 in steps of 0.1 for $E_{U}$ discovery recall. We optimize $d_{m}$ in range 0.5 to 1.0, in steps of 0.0001 for NMI. %
We optimize $d_{e}$ for $E_{U}$ discovery recall in range 50 to 250 in steps of 10. For results, see Table \ref{tab:adaptation}.

We report recall of 81\% and precision of 6\% for clusters referring to unknown entities. Recall of clusters referring to known entities is 88\% with precision 96\%. Clustering NMI is 0.92.

\begin{table}[]
    \centering
    \footnotesize{
    \begin{tabular}{cl }
    \toprule
      \bf Parameter & \bf Value \\
     \midrule
      $d_{m}$ &  0.8171 \\
      $s_{m}$ & 0.5 \\
      $d_{e}$ & 110  \\
    \bottomrule
    \end{tabular}
    \caption{Hyper-parameters adaptation phase}
    \label{tab:adaptation}}
\end{table}

\section{Effect of re-training}
We show that by re-training L after indexing, L learns to circumvent $E_{U}$: We identify known entities part of the training set that are in close proximity of unknown entities ({\it confusable known entities}). We compare the average similarity between mentions and their respective linked entity when adding unknown entities before training vs. after training. Mean similarity when adding unknown entities before training is 93.28 for confusable known entities and 92.57 for other known entities. A t-test shows that this difference is significant (p-value of 0.0001 with $<$ 0.05). As a reference, mean similarity when adding unknown entities post training is 92.65 irrespective of whether they are confusable or not.

\section{Disambiguation Results}
Besides end-to-end performance, we also report entity disambiguation performance with oracle mention detection in Table \ref{tab:disambiguation}.

\section{Wikipedia Results}

We report performance on Wikipedia $D_{t2}$-test in Table \ref{tab:e2ewiki}. Due to a smaller $D_{adapt}$, end-to-end performance is lower. When filtering Wikipedia $D_{t2}$-test for mentions of discovered entities, $L_{t1}$-Cluster-Gold precision is 40.5 and $L_{t1}$-Cluster recall is 15.3.

\begin{table*}[htp]
    \centering
    \begin{tabular}{lccccccccc}
    \toprule
    &\multicolumn{6}{c}{\bf OSCAR}\\
    
    &\multicolumn{3}{c}{\bf Unknown Entities} & \multicolumn{4}{c}{\bf Known Entities}\\
    
     \bf Model & \bf R@1 & \bf P@1 & \bf NMI  & \bf R@1 & \bf P@1 & \bf NMI \\
    \midrule[0.01pt]
    $L_{Dt1} $ & 0.0 & 0.0 & 0.0  & 92.2 & 92.2 & 96.0 \\
    $L_{Dt2}$  & 63.5 & 45.3 & 96.8 & 90.0 & 90.2 & 96.0 & \\
    \midrule
    $L_{t1}$-Descp & 58.0 & 33.9 & 96.3 & 92.1 & 92.3 & 96.1 \\
    \midrule[0.01pt]
    $L_{t1}$-Mention & 26.2 & 30.8 & 92.7 & 92.2 & 92.2 & 96.0  \\
    \midrule[0.01pt]
    EDIN ($L_{t1}$-Cluster) & 27.9 & 34.1 & 93.4 & 92.2 & 92.2 &  96.2 \\
    \bottomrule
    \end{tabular}
    \caption{\textbf{Entity Disambiguation performance } on  OSCAR $D_{t2}$-test.}
    \label{tab:disambiguation}
\end{table*}

\begin{table*}[!htbp]
    \centering
    \begin{tabular}{lccc|cccccc}
    \toprule
    &\multicolumn{3}{c}{\bf Unknown Entities} & \multicolumn{3}{c}{\bf Known Entities} \\
    \toprule
     \bf Model & \bf R@1 & \bf P@1 & \bf NMI & \bf R@1 & \bf P@1 & \bf NMI \\
    \midrule[0.01pt]
    $L_{t1} $ & 0.0 & 0.0 & 0.0 & 70.5 & 75.8 & 95.4 \\
    $L_{t2}$  & 33.6 & 25.0 & 98.3 & 70.6 & 75.4 & 95.3 \\
    \midrule
    $L_{t1}$-Descp & 33.9 & 20.0 & 98.0 & 71.2 & 74.4 & 95.3 \\
    \midrule[0.01pt]
    $L_{t1}$-Cluster-Gold & 7.8 & 55.6 & 90.6 & 70.1 & 75.9 & 95.6  \\
    EDIN ($L_{t1}$-Cluster) & 1.8 & 15.4 & 93.4 & 71.1 & 74.1 & 95.3 \\
    \bottomrule
    \end{tabular}
    \caption{\textbf{End-to-end EL performance} on Wikipedia $D_{t2}$-test.}
    \label{tab:e2ewiki}
\end{table*}

\end{document}